\documentclass{article}

\usepackage[preprint]{neurips_2023}

\usepackage[utf8]{inputenc} % allow utf-8 input
\usepackage[T1]{fontenc}    % use 8-bit T1 fonts
\usepackage{hyperref}       % hyperlinks
\usepackage{url}            % simple URL typesetting
\usepackage{booktabs}       % professional-quality tables
\usepackage{amsfonts}       % blackboard math symbols
\usepackage{nicefrac}       % compact symbols for 1/2, etc.
\usepackage{microtype}      % microtypography
\usepackage{xcolor}         % colors
\usepackage{graphicx}
\usepackage[font=small,labelfont=bf]{caption}
\usepackage{wrapfig}
\usepackage{adjustbox}

\newcommand{\jack}[1]{\textcolor{cyan}{\bf [jack: #1]}}

\title{Novice Learner and Expert Tutor:\\
Evaluating Math Reasoning Abilities of \\
Large Language Models with Misconceptions}

\author{%
  Naiming Liu \\
  Rice University\\
  \texttt{nl35@rice.edu} \\
  \And
  Shashank Sonkar \\
  Rice University\\
  \texttt{ss164@rice.edu} \\
  \And
  Zichao Wang \\
  Adobe Research \\
  \texttt{jackwa@adobe.com} \\
  \AND
  Simon Woodhead \\
  Eedi \\
  \texttt{simon.woodhead@eedi.co.uk} \\
  \And
  Richard G. Baraniuk \\
  Rice University \\
  \texttt{richb@rice.edu} \\
}

\begin{document}

\maketitle

\begin{abstract}
We propose novel evaluations for mathematical reasoning capabilities of Large Language Models (LLMs) based on mathematical misconceptions. 
Our primary approach is to simulate LLMs as a novice learner and an expert tutor, aiming to identify the {\emph incorrect} answer to math question resulted from a specific misconception and to recognize the {\it misconception(s)} behind an incorrect answer, respectively. 
Contrary to traditional LLMs-based mathematical evaluations that focus on answering math questions correctly, our approach takes inspirations from principles in educational learning sciences. We explicitly ask LLMs to mimic a {\it novice learner} by answering questions in a specific incorrect manner based on incomplete knowledge; and to mimic an {\it expert tutor} by identifying misconception(s) corresponding to an incorrect answer to a question.
Using simple grade-school math problems, our experiments reveal that, while LLMs can easily answer these questions correctly, they struggle to identify 1) the incorrect answer corresponding to specific incomplete knowledge (misconceptions); 2) the misconceptions that explain particular incorrect answers. 
Our study indicates new opportunities for enhancing LLMs' math reasoning capabilities, especially on developing robust student simulation and expert tutoring models in the educational applications such as intelligent tutoring systems.
\end{abstract}

\section{Introduction}

The fast-evolving development in math problem solving capabilities of Large Language Models (LLMs) demands appropriate evaluations to measure progress and identify opportunities for further improvements.
Most, if not all, existing evaluation approaches compute the percentage of math questions that LLMs can {\it correctly solve}, given a set of such questions. For instance, the state-of-the-art LLM, GPT-4~\citep{gpt4} has achieved a commendable performance on benchmark datasets, including GSM8K (grade-school math word problems)~\citep{gsm8k}, MATH (competition-level complex problems)~\citep{mathdataset}, and NumGLUE (numerical reasoning)~\citep{numglue}.  

In this paper, we ask a different question: how do LLMs perform if we instruct them to complete a math problem {\em incorrectly}, and, moreover, not just incorrectly in any arbitrary way but {\em in a specific manner}? On the flip side, if given a specifically incorrect answer to a math problem, can the LLMs detect and analyze the specific error in the answer?
Analyzing LLMs' performance in this setting can reveal if LLMs truly understand the mathematical concepts required to answer a particular math question, enhance the explainability of their mathematical problem-solving behaviors, and suggest a way to evaluate their math reasoning robustness in real-world use cases such as education. 
Our new approach of evaluating LLMs can also be seen as a special instance of counterfactual reasoning \citep{dupe}, in which LLMs will need to abide to new knowledge and discard the information it has already encoded. Our investigation also has far-reaching potential implications for intelligent tutoring systems in education~\citep{sonkar2023code, sonkar2023class} with an emphasis on correctly understanding student's error and misconceptions for effective guidance.

\subsection{Contributions}
% \jack{we can probably either remove this subsection or integrate it together with the beginning of the intro}
% \jack{or maybe shorten the 3rd and 4th paragraphs in the intro}

In this work, we propose "novice learner and expert tutor" approach to evaluate the mathematical reasoning capabilities of LLMs based on math misconceptions. 
Instead of solely focusing on the LLMs' success rate in answering math questions correctly, we measure whether LLMs can identify the wrong answer to a question corresponding to a specific set of math misconceptions, simulating a novice learner, and whether LLMs can identify the set of misconceptions that can explain a particular wrong answer to a question. Such investigation provides a new way to evaluate LLMs' capabilities in mathematical reasoning and has far-reaching implications for education applications. We conduct experiments on grade-school math problems and demonstrate that LLMs face challenges in replicating the behaviors of 1)  novice learners who make mistakes, and 2) expert tutors who precisely understand why learners make certain mistakes.

\subsection{Problem formulation}

We define a natural language representation of a mathematical misconception $m$ as an element in a set of misconceptions $\mathcal{M}$. 
% Each novice learner, represented by $i$, has a subset of misconceptions  $\mathcal{M}_i\subset \mathcal{M}$. 
For each multiple choice question $q$, we denote its four possible answer choices as $\mathcal{A}_q$ where the correct answer is represented as $a_q$ and the incorrect answer choice caused by the misconception $m$ as $a_{q, m}$. For simplicity, we assume each incorrect answer choice is determined by a single misconception. 

For ``novice learners'' evaluation, we obtain the LLM $f_\theta$'s output answer to question $q$ based on the specific misconceptions: $\hat{a}_{q,m} = f_\theta (p(q, m, \mathcal{A}_q))$
% \begin{align*}
%     \hat{a}_{q,m} = f_\theta (p(q, m, \mathcal{A}_q))
% \end{align*}
% $\hat{a}_{q,m} = f_\theta (p(q, m, \mathcal{A}_q)$, 
where $p(\cdot)$ is the prompt. We then compute the accuracy $s_1 = \frac{1}{N}\sum_{i=1}^N {1} \{a_{q,m} = \hat{a}_{q,m} \}$. 

For the "expert tutors" evaluation, we derive LLM $f_\theta$'s output misconception given a question, a particular incorrect answer and a subset of misconceptions $\mathcal{M}_n \subset \mathcal{M}$:
% \begin{align*}
%     \hat{m} = f_\theta (p(q, a_{q,m}, \mathcal{M}_n))
% \end{align*}
$\hat{m} = f_\theta (p(q, a_{q,m}, \mathcal{M}_n))$, 
where $p(\cdot)$ is the prompt. We compute the accuracy as $s_2 = \frac{1}{N}\sum_{i=1}^N {1} \{m_i = \hat{m}_i \}$.

\subsection{Dataset}
\label{sec:data-analysis}

\begin{wraptable}{R}{0.5\textwidth}
\centering
\begin{tabular}{lll}
\toprule
& \multicolumn{1}{c}{\textbf{w/o exp}} & \multicolumn{1}{c}{\textbf{with exp}} \\
\midrule
\textbf{GPT-3.5} & 0.576 $\pm$ 0.025 &  0.805 $\pm$ 0.027 \\
\textbf{GPT-4} & 0.774 $\pm$ 0.019 &  \textbf{0.948 $\pm$ 0.009} \\
\bottomrule
\end{tabular}
\caption{Accuracy of both GPT-3.5 and GPT-4 when answering math questions in our dataset.}
\label{tab:accuracy}
\end{wraptable} 

Our research utilizes a dataset from Eedi's platform\footnote{\url{https://eedi.com/home}}, which comprises multiple-choice grade-school level math questions. More details of the dataset and preprocessing steps are in Appendix~\ref{app:dataset}. We explore \textbf{two prompting strategies} to examine the accuracy of LLMs to {\em correctly} answer the questions. 
The \textbf{``w/o exp"} prompting strategy provides LLMs with a question and four answer choices, while requesting correct answer as the output. The \textbf{``with exp"} prompt demands the LLMs to justify the answer with explanation \textbf{prior to} providing an answer itself, which is similar to the Chain-of-Thoughts prompting~\citep{CoT}. Table~\ref{tab:accuracy} provides the accuracy of question answering, where GPT-4 could achieve a near-perfect accuracy of  \textbf{94.8\%}. 
It clearly demonstrates that LLMs can {\em accurately} solving the math questions in the dataset, when providing with appropriate prompts. It solidifies the baseline performance of LLMs with our dataset, which offers potential for further exploration of LLM's ability to answer math questions {\em incorrectly in a specific manner}. 
% An example of the incorrectly answered question by GPT-4 explanations can be found in Figure~\ref{fig:incorrect}.

\begin{figure}[t!]
    \centering
    \includegraphics[width=\linewidth]{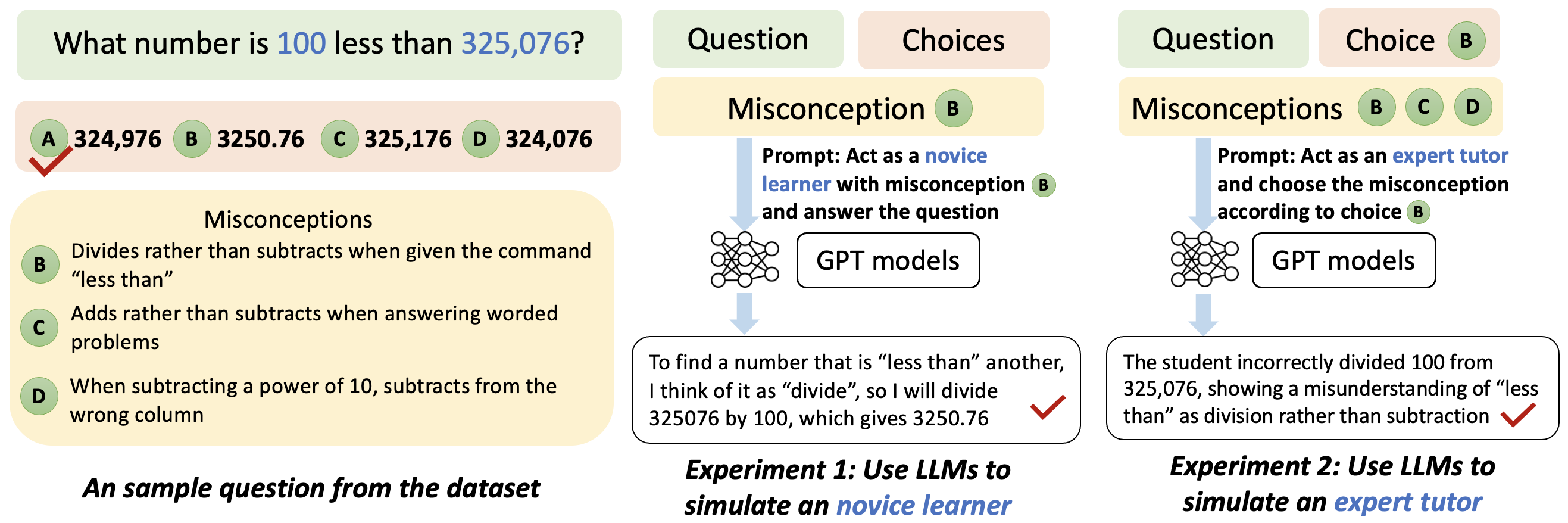}
    % \vspace{-10pt}
    \caption{(\textbf{Left}) An example question from the dataset with question, choices and the corresponding misconceptions. (\textbf{Middle}) A depiction of experiment 1 in which we use LLMs to mimic the role of novice learners. (\textbf{Right)} An illustration of experiment 2 where we use LLMs to simulate expert tutors. }
    \label{fig:exp}
\end{figure}

\section{Experiments}

We conduct two experiments to evaluate if LLMs can effectively simulate novice learners and expert tutors, as shown in Figure~\ref{fig:exp}. We use \textbf{with exp} prompting strategy mentioned in Section~\ref{sec:data-analysis} unless defined otherwise. We utilize GPT-3.5 and GPT-4~\citep{gpt4} by OpenAI, with a temperature of 0.9. To automate the evaluation process, we prompt the LLMs to deliver structured responses in a predefined json format: \{ {\it "Explanation": "...", "Choices": "A/B/C/D" }  \}. We ensure statistical significance by reporting all results based on the mean and standard deviation derived from four random, independent trials.

\subsection{Experiment 1: Using LLMs to simulate a novice learner}
\label{sec:exp-1}

Our first experiment was designed to explore the potential of LLMs to mimic an novice learner by answering math questions incorrectly in a particular pattern. More specifically, we present LLMs with a datapoint $(q, m, \mathcal{A}_q)$.
Instead of seeking a correct answer, we require the LLMs to provide the specific incorrect answer $a_{q,m}$ that aligned with the provided math misconception $m$ because our primary objective is to assess LLMs' potential to distinguish predictable error patterns based on incomplete knowledge (misconception). In order to conduct a comprehensive experiment, we explore four prompt settings as follows: (1). [\textbf{zero (w/o exp)}]: zero-shot prompting without asking for explanations; (2). [\textbf{zero (with exp)}]: zero-shot prompting requires explanations; (3). [\textbf{few (rand)}]: few-shot prompting uses 2 \textbf{random examples} and requires explanations; (4). [\textbf{few (same)}]: few-shot prompting uses 2 \textbf{examples of the same misconceptions} as the given misconception and requires explanation (only a subset of 152 datapoint is suitable)

% \begin{itemize}
% \small
% \itemsep0em 
%     \item \textbf{p1}[\textbf{zero (w/o exp)}]: zero-shot prompting without asking for explanations
%     \item \textbf{p2} [\textbf{zero (with exp)}]: zero-shot prompting requires explanations
%     \item \textbf{p3} [\textbf{few (rand)}]: few-shot prompting uses 2 \textbf{random examples} and requires explanations    
%     \item \textbf{p4} [\textbf{few (same)}]: few-shot prompting uses 2 \textbf{examples of the same misconceptions} as the given misconception and requires explanation (only a subset of 152 datapoint is suitable)
% \end{itemize}

We provide the prompts for \textbf{zero (with exp)} below and the remaining prompts in Appendix~\ref{app:prompts}.

\begin{center}
\fbox{\begin{minipage}{\dimexpr\textwidth-1cm}
\small
\texttt{You are a student with some math misconceptions. \\
Your task is to answer a multiple-choice math question as a student who doesn't understand the concepts completely and has some math misconceptions. Your answer should be based on the given misconception. DO NOT just give the correct answer! Also provide some explanations on why you choose your answer.
 }
\end{minipage}}
\end{center}

Table~\ref{tab:exp1} provides the results for each prompting strategy. 
The results reveal that an effective zero-shot prompting strategy can yield an accuracy of 61.7\% for GPT-4 model to select the ``appropriate" incorrect choice based on a specific math misconception. While the accuracy significantly exceeds the rate of random selection, the performance is still not quite satisfactory when compared to GPT-4's near-perfect performance at answering the questions accurately. 
The few-shot prompting strategy with randomly selected examples leads to a even worse performance than zero-shot prompting. Our hypothesis is that the randomly selected examples might not always be relevant or precise, potentially introducing more distractions and ``noise" in the learning processing. 
% Furthermore, utilizing the identical set of \textbf{p4}'s 152 datapoint in the setting of \textbf{p3} leads to an accuracy of 46.6\% for GPT-3.5 and 64.1\% for GPT-4. 
Furthermore, the relatively higher accuracy of few-shot prompting with same misconceptions suggest that appropriate few-shot examples can ultimately enhance performance. In the future work, we aim to extend our study with in-context learning to choose the most suitable examples.

Overall, although LLMs demonstrate impressive accuracy when providing accurate answers to grade-school math questions, their ability to act as novice learners and answer the question {\em incorrectly in a specific way} remains limited. There's still considerable potential for improving the way LLMs ``controllably" generate specified erroneous responses to math problems following particular instructions. 

\begin{table}[t!]
\begin{minipage}[h]{0.49\textwidth}
\small
\centering
% \begin{adjustbox}{max width=\linewidth}
\begin{tabular}{lll}
\toprule
& \multicolumn{1}{c}{\textbf{zero (w/o exp)}} & \multicolumn{1}{c}{\textbf{zero (with exp)}} \\
\midrule
\textbf{GPT-3.5} & 0.371 $\pm$ 0.010 & 0.425 $\pm$ 0.011 \\
\textbf{GPT-4} & 0.534 $\pm$ 0.007 & 0.617 $\pm$ 0.005 \\
\midrule
& \multicolumn{1}{c}{\textbf{few (rand)}} & \multicolumn{1}{c}{\textbf{few (same)}}  \\
\midrule
\textbf{GPT-3.5} & 0.415 $\pm$ 0.023 & 0.497 $\pm$ 0.031 \\
\textbf{GPT-4} & 0.600 $\pm$ 0.014 & 0.679 $\pm$ 0.020 \\
\bottomrule
\end{tabular}
% \end{adjustbox}
\vspace{5mm}
\caption{Accuracy of evaluating GPT-3.5 and GPT-4 models' capability to behave as a novice learner. We use four prompting strategies. Compare to results in Table 1, LLMs perform considerably worse in simulating novice learners with incomplete knowledge.}
\label{tab:exp1}
\end{minipage}
% \hspace{-2mm}
\hfill
\begin{minipage}[h]{.49\textwidth}
\centering
% \begin{adjustbox}{max width=\linewidth} \vspace{0pt}
\includegraphics[width=\linewidth]{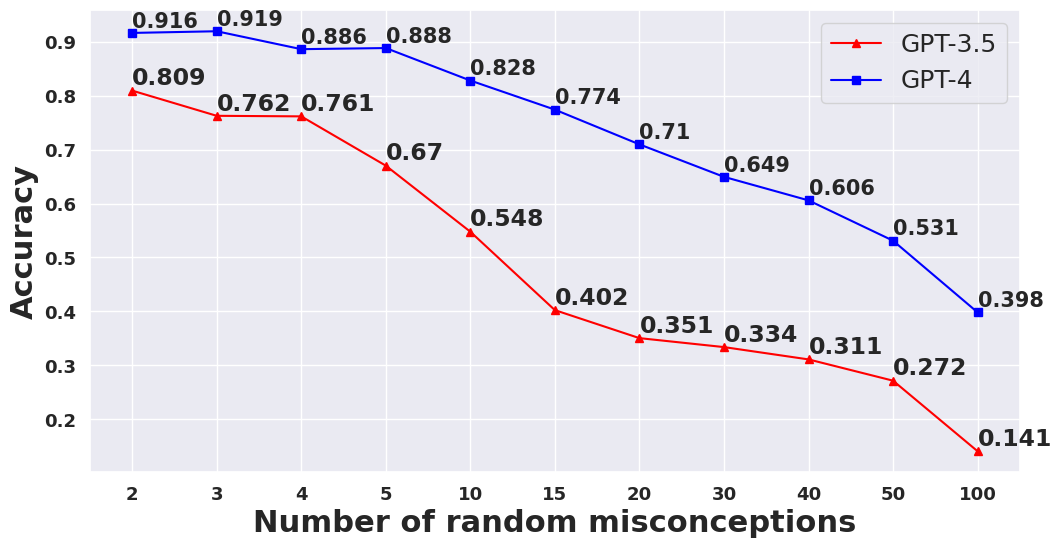}
% \end{adjustbox}
\vspace{1pt}
\captionof{figure}{Performance of GPT models mimicking expert tutors. Increasing the number of misconceptions results in a decline of accuracy in detecting the misconception that explains a wrong answer.}
\label{fig:exp2}
\end{minipage}
\end{table}

\subsection{Experiment 2: Using LLMs to simulate an expert tutor}

We explore the capability of LLMs to simulate an expert tutor, particularly on pinpointing misconception(s) specific to the incorrect student answer of a math question. We provide the LLMs with a datapoint $(q, a_{q,m}, \mathcal{M}_n)$, where $\mathcal{M}_n$ contains a sequence of $n$ numbers of misconceptions. Among these misconceptions, a specific misconception $m$ corresponds to the incorrect answer $a_{q,m}$ while the rest of $n-1$ misconceptions are randomly selected from the misconceptions pool $\mathcal{M}$. We then prompt the LLMs to identify the correct misconception $m$. Additionally, we incrementally increase the number of misconceptions presented to the LLMs. We use the following prompt to guide the LLMs:

\begin{center}
\fbox{\begin{minipage}{\dimexpr\textwidth-1cm}
\small
\texttt{You are an expert math tutor who knows about all grade-school level math misconceptions. \\
Your task is to select the accurate type of misconceptions your student have based on the (incorrect) answer he/she gives to a multiple-choice math question. You will be given $n$ misconception types. Your selected misconception type should correspond to the given question and answer. You should also provide some explanations on why you choose the misconception.}
\end{minipage}}
\end{center}

Our results in Figure~\ref{fig:exp2} reveal that with a starting point of $n=4$ (one real and three random) misconceptions, GPT-4 achieves an impressive accuracy of 91.9\%. It demonstrates that LLMs, when provided with a limited amount of misconceptions, can effectively mirror a tutor’s ability to identify the correct misconceptions causing inaccurate answer. However, an inversely proportional relationship between the number of misconceptions and the performance of both LLMs is observed as we increase the number of misconceptions. The performance drops down to an average of 39.8\% for GPT-4 as the misconception number increases to 100. The ideal proficiency level of an expert tutor would involve the recognition and differentiation of at least 100 types of grade-school level math misconceptions effortlessly. Judging from this stringent criterion, LLMs still have much room for improvement in approximating the comprehensive knowledge and adaptive problem-solving acumen of an expert tutor.

\section{Conclusion}
We propose novel evaluations of the mathematical reasoning capabilities of LLMs based on math misconceptions. 
Our experiments indicate that LLMs struggle to simulate ``novice'' learners with specific knowledge gaps. We also find that even though LLMs are capable of emulating a ``tutor'', they could not simulate the proficiency of an ``expert'' tutor, even on simple grade-school level questions.
Our evaluations reveal opportunities to further improve mathematical question answering capabilities of LLMs to additionally consider scenarios where incomplete knowledge is assumed and misconceptions are taken into account. Doing so will help probe whether LLMs truly understand the mathematical concepts, improve the LLMs' explainability when performing math question answering, and unlock their potential use in improving intelligent tutoring systems in education use cases.
Avenues of future work include 1) expanding our current evaluation to additional scenarios and LLMs, such as LLaMA-2~\citep{touvron2023llama} or Vicuna~\citep{vicuna2023}; and 2) improving LLMs' capabilities in handling and understanding misconceptions.

\iffalse
There are some avenues for our future directions. First, we plan to combine several questions as a "math quiz" to LLMs and enhance the complexity of our evaluations of misconception. Second, we aim to incorporate other open-source LLMs, such as Llama 2~\citep{touvron2023llama} or Vicuna~\citep{vicuna2023}, in our evaluations. Apart from prompting, we will fine-tune these models to explore if we could increase the performance on both the tasks. Third, we plan to gather more data on mathematical misconceptions and provide more clear definition and classifications to the misconceptions, such as formulate them as counter-factual reasoning statements. 
\fi

% \nl{future work: 0. increase the number of questions introduced and make it like a "quiz". 1. fine-tune smaller LMs such as LLama-2? 2. clearly define misconceptions and formulate it as a counter-factual reasoning? 3. in-context learning ?  what else???????}

\bibliography{ref}
\bibliographystyle{acl_natbib}

% \newpage
\appendix

\section{Dataset Statistics}
\label{app:dataset}

\begin{figure}[t!]
    \centering
    \includegraphics[width=\linewidth]{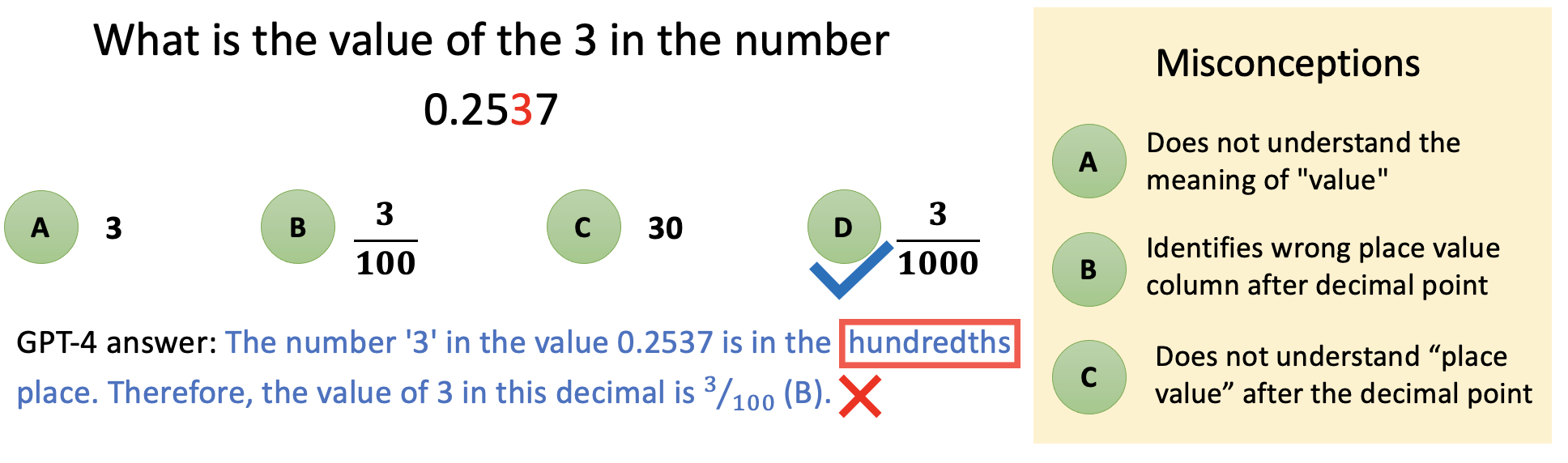}
    % \vspace{-10pt}
    \caption{An sample question from our dataset along with associated misconceptions for each incorrect choice. It also includes an erroneous response given by GPT-4.}
    \label{fig:incorrect}
\end{figure}

Our Eedi dataset includes multiple-choice grade-school level math questions. Each question contains four answer choices, of which only one is correct. Additionally, we have a list of expert-curated student misconceptions corresponding to the incorrect choices, though not every choice possesses an associated misconception.
We preprocess the dataset using the two following steps: 1). remove all the questions that contain images or graphs; 2). remove all the questions whose different choices reveal identical misconceptions. 
After the preprocessing steps, the dataset contains 213 questions, paired with 482 choices that revealed a collective total of 319 student misconceptions.
An example of a math question, the corresponding misconceptions and an instance of GPT-4's incorrect answer to the question is provided in Figure~\ref{fig:incorrect}.

\section{Prompts}
\label{app:prompts}

Prompt for LLMs to correctly answer the questions (\textbf{w/o exp}): 
\begin{center}
\fbox{\begin{minipage}{\dimexpr\textwidth-1cm}
\small
\texttt{Your are a helpful assistant for answering simple math questions. \\
Your task is to give the correct answer of the provided math multiple choice questions. Put your answer as the letter index (A/B/C/D) in json format as below. \\
\{"Answer": "A/B/C/D" \}}
\end{minipage}}
\end{center}

Prompt for LLMs to correctly answer the questions (\textbf{w/o exp}): 

\begin{center}
\fbox{\begin{minipage}{\dimexpr\textwidth-1cm}
\small
\texttt{Your are a helpful assistant for answering simple math questions. \\
Your task is to give the correct answer of the provided math multiple choice questions. Also provide explanations on why you choose the answer. Put your answer as the letter index (A/B/C/D) in json format as below. \\
\{"Explanation": "...", "Answer": "A/B/C/D" \}}
\end{minipage}}
\end{center}

Prompt for LLMs to act as novice learners (\textbf{zero (w/o exp)}): 

\begin{center}
\fbox{\begin{minipage}{\dimexpr\textwidth-1cm}
\small
\texttt{You are a student with some math misconceptions. \\
Your task is to answer a multiple-choice math questions as a student who doesn't understand the concepts completely and has some math misconceptions. Your answer should be based on the given misconception. Choose a choice from the four answer choices provided and DO NOT just give the correct answer! Put your answer (A/B/C/D) in json format with your choice as shown below. \\
\{"Choice": "A/B/C/D" \}
}
\end{minipage}}
\end{center}

Prompt for LLMs to act as novice learners (\textbf{few (rand / same}): 

\begin{center}
\fbox{\begin{minipage}{\dimexpr\textwidth-1cm}
\small
\texttt{You are a student with some math misconceptions. \\
Your task is to answer a multiple-choice math questions as a student who doesn't understand the concepts completely and has some misconceptions. Some examples of questions, misconceptions and answer triples are provided below: \\
    \{few-shot examples...\} \\
Your answer should be based on the given misconception and DO NOT just give the correct answer! You should also provide some explanations on why you choose the choice. Put your answer in json format with your choice and explanation as shown below. \\
\{"Explanation": "...", "Answer": "A/B/C/D" \}
}
\end{minipage}}
\end{center}

%%%%%%%%%%%%%%%%%%%%%%%%%%%%%%%%%%%%%%%%%%%%%%%%%%%%%%%%%%%%

\end{document}